\documentclass{svproc}
\usepackage{url}
\usepackage{graphicx}
\usepackage{caption}
\usepackage{dblfloatfix} 
\usepackage[colorlinks=true,
            linkcolor=red,
            urlcolor=blue,
            citecolor=blue]{hyperref}
\usepackage{subcaption}
\usepackage{comment}

\begin{document}
\mainmatter              
\title{From Videos to URLs: A Multi-Browser Guide To Extract User's Behavior with Optical Character Recognition}
\titlerunning{Video to URLs with OCR}  
%

\author{Mojtaba Heidarysafa\inst{1},James Reed\inst{2}, Kamran Kowsari\inst{1}, April Celeste R. Leviton\inst{2,5}, Janet I. Warren\inst{2,4}, \and Donald E. Brown\inst{1,3}
}
\authorrunning{Heidarysafa et al.} 
%
\tocauthor{Ivar Ekeland, Roger Temam, Jeffrey Dean, David Grove,
Craig Chambers, Kim B. Bruce, and Elisa Bertino}
\institute{Department of Systems \& Information Engineering, University of Virginia,  VA
\and
Institute of Law, Psychiatry, \& Public Policy, University of Virginia,  VA
\and
Data Science Institute, University of Virginia,  VA
\and
Department of Psychiatry \& Neurobehavioral Sciences, University of Virginia,  VA
\and
Department of Sociology, University of California, Riverside, CA
}

\maketitle              

\begin{abstract}
Tracking users' activities on the World Wide Web (WWW) allows researchers to analyze each user's internet behavior as time passes and for the amount of time spent on a particular domain. This analysis can be used in research design, as researchers may access to their participant's behaviors while browsing the web. Web search behavior has been a subject of interest because of its real-world applications in  marketing, digital advertisement, and identifying potential threats online. In this paper, we present an image-processing based method to extract domains which are visited by a participant over multiple browsers during a lab session. This method could provide another way to collect users' activities during an online session given that the session recorder collected the data. The method can also be used to collect the textual content of web-pages that an individual visits for later analysis\footnote{Code is shared as an open source tool at \url{https://github.com/mojtaba-Hsafa/OCR-browser-domain-extractor}\\This research was funded by Grant 2016-ZA-BX-K002 from the National Institute of Justice, Office of Justice Programs, US Department of Justice, to the University of Virginia.}.
\keywords{Web search, User behavior, Image processing, Optical character recognition}
\end{abstract}
\section{Introduction}

Since the invention of World Wide Web (WWW) in the 1980s by Tim Berners-Lee, the internet has continued to impact our society, culture, and everyday life activities. Given the explosive increase in information on internet, it has become the first resource people turn to when seeking information. One side effect of using the web for information retrieval is that looking at users' behavior patterns on the internet opens up doors to understanding their interests. Researchers work to leverage this insight to improve multiple facets of user experiences over the web. Areas such as designing interfaces, marketing, and digital advertisement are directly benefited from such research.

Furthermore, it is also of interest to understand how people surf the internet and the pathways that lead them to specific places. One such places is the dark web where illegal activities including promoting terrorism and other cyber crimes have been spreading~\cite{spalevic2017use}. In order to use the dark web, Tor, a spacial browser is needed which provides access to dark web content anonymously using multiple nodes which reroute the connection. In our study, the usual scenario of information retrieval via the internet and dark web consists of the user beginning their search by accessing a regular internet browser,  and then switching to the dark web using Tor. This process may repeat multiple times until the user finds the information they are seeking. Understanding the time spent on each domain in this circumstance is not a trivial task. However, given that this is a research experiment, one could use videos of these online search session which recorded the whole screen during the interaction. To accomplish this, utilized a general framework that took a stream of screen images and outputs the domain or URL visited in that frame regardless of the type of browser (internet or Tor). These URLs were collected for each user and are able to be analyzed separately.

The purpose of this paper is to demonstrate an image-based approach that can be used in other scenarios where multiple browsers are working in parallel, given that the computer screen has been recorded. Such conditions might be suitable for research conducted in controlled environments, such as online searching sessions in a lab. Moreover, since the screen has been recorded, other analysis can be done using the same approach. As an example, one can analyze the textual content of web pages by using optical character recognition and the technique provided in this paper. The method that is described in this paper is fully open-source and available to use for similar tasks. 

This paper is structured as follows. In Section~\ref{Sec.2} we discuss related work  for tracking users' activities on the web as well as image processing techniques specifically, optical character recognition (ORC). Section~\ref{Sec.3} describes our implementation in details. Subsequently, in Section~\ref{Sec.4}, we present the result of this approach. Finally,  in Section~\ref{Sec.5} we discuss possible improvement to this approach and provide our concluding remarks.
\section{Related Work}\label{Sec.2}
As mentioned before, users' interactions with materials on the internet reveal details related to the individuals' interests.
Since the early days of internet, researchers have been using this insight to try to understand online user behaviour.~\cite{cat:pit}. Discovering user patterns by mining web usage behaviors has been addressed by Srivastava, et al~\cite{Sirv:cool:Desh}. Specifically, web search patterns have been investigated extensively by researchers~\cite{hsieh}~\cite{rose:levi}~\cite{Hols:Str}. The result of this scholarship has provided ways for improving search engine rankings, advertisement placements, amd search engine performance~\cite{agi:bril}~\cite{bor:mark}. At an upper level, this research  relies on log-files of user interactions which are gathered from the client or server side. Usual approaches include the use of add-ons such as Firefox Slogger or d{\'e}j{\`a}click (which rely on the browser history log files) or standalone software such as Track4Win which tracks internet usage and computer activities. Although these tools might provide insights of user activities, they cannot be used in every situation we might encounter during experiments with users. As an example, most add-ons are specific for a browser and a different browser, like Tor, would not allow add-on installation. Similarly as its name suggests, Track4Win is mostly suitable for machines with Windows as their operating system. Alternatively however, one can design the experiment such that they record the interaction of users with Internet using a video recorder such as OBS studio for windows machines or built-in quick-time screen recorders. In such scenarios,  these videos combined with image processing tools and Optical Character Recognition (OCR) techniques can work to extract user interactions with internet, regardless of the browser and operating systems. \\
\begin{figure}[tp]
\centering
\includegraphics[width=\textwidth]{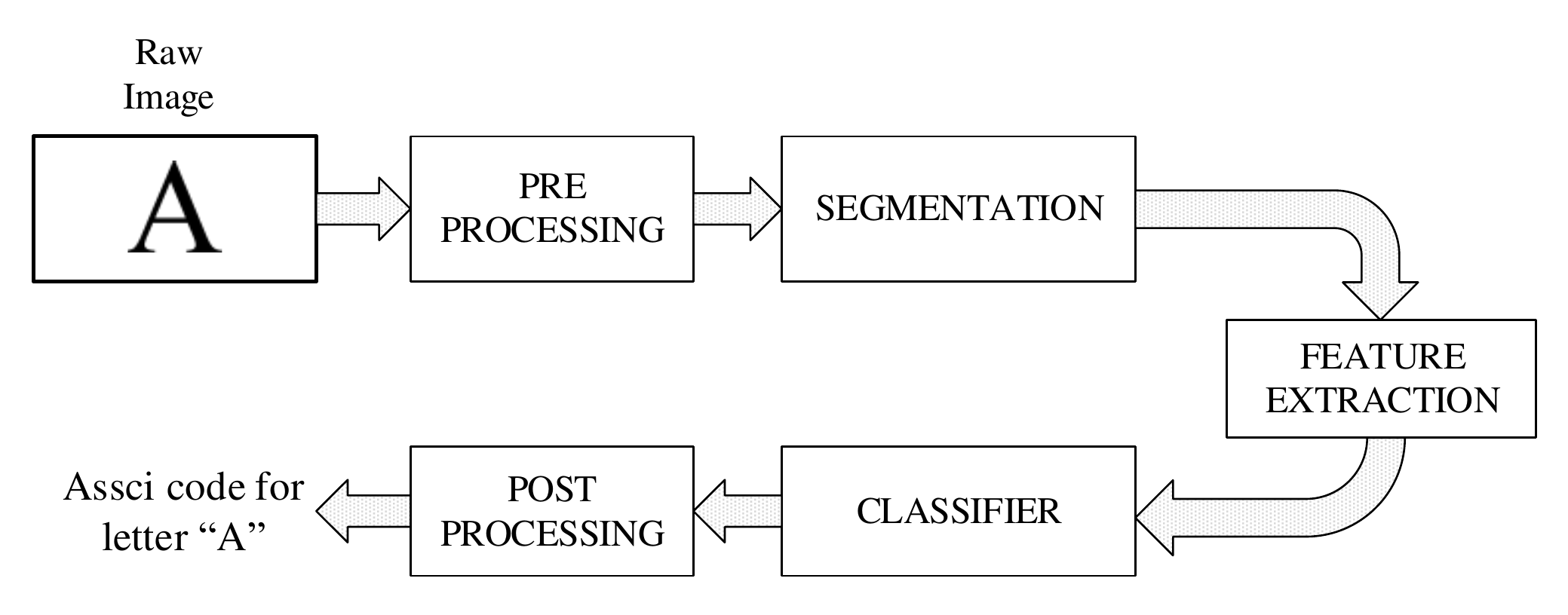}
\caption{Structure of OCR system}
\vspace{-20pt}
\label{ocr_struct}
\end{figure}

A major cornerstone of the previously mentioned approach is  Optical Character Recognition.  OCR is used in many practical applications such as scanners in ATMs and office scanning machines which use OCR to understand characters~\cite{mori:nish}. OCR will be referred to either as off-line (where the writing or printing is completed) or as on-line (where character recognition will be performed simultaneously with writing). Different tasks such as hand-writing recognition or hand-written script verification may be performed with OCR techniques. However in this work, we focused on character recognition in a printed text as it is in url field of a browser.\\

Figure~\ref{ocr_struct}  shows the structure of an OCR system. Pre-processing usually includes steps such as binarization, noise removal, and skew detection~\cite{Chandarana}. Next, segmentation will be performed to deconstruct an image into lines or characters. Feature extraction is another important step in OCR structure and different approaches have been suggested to perform this task~\cite{Kumar:Bhatia}. Finally the most important step is classification of the character with high accuracy. Traditionally, this has been done by template matching or correlation-based techniques~\cite{Chandarana}. Other researchers have presented techniques such as fast tree-based clustering,  HMM based on combination of frequency and time domain, and K- Nearest Neighbours for classification task~\cite{Berchmans:Deepa}. Moreover, other machine learning classifiers for this step have also gained a lot of attention.  In recent years, Support Vector Machines (SVM) have been used as powerful classifiers at the last part of an OCR system~\cite{Yafang}. Also, Artificial Neural Network (ANN) has been used due to its high tolerance for noise provided correct features~\cite{Sameeksha}.
As one can see, we used tesseract OCR which uses a two-step process for classification with an adaptive classifier to perform the recognition.

We will describe the structure of this open-source OCR along with other necessary steps such as image pre-processing and target selection with template matching in the next section.

\vspace{-.51cm}
\begin{figure}[t]
\centering
\includegraphics[width= 10cm]{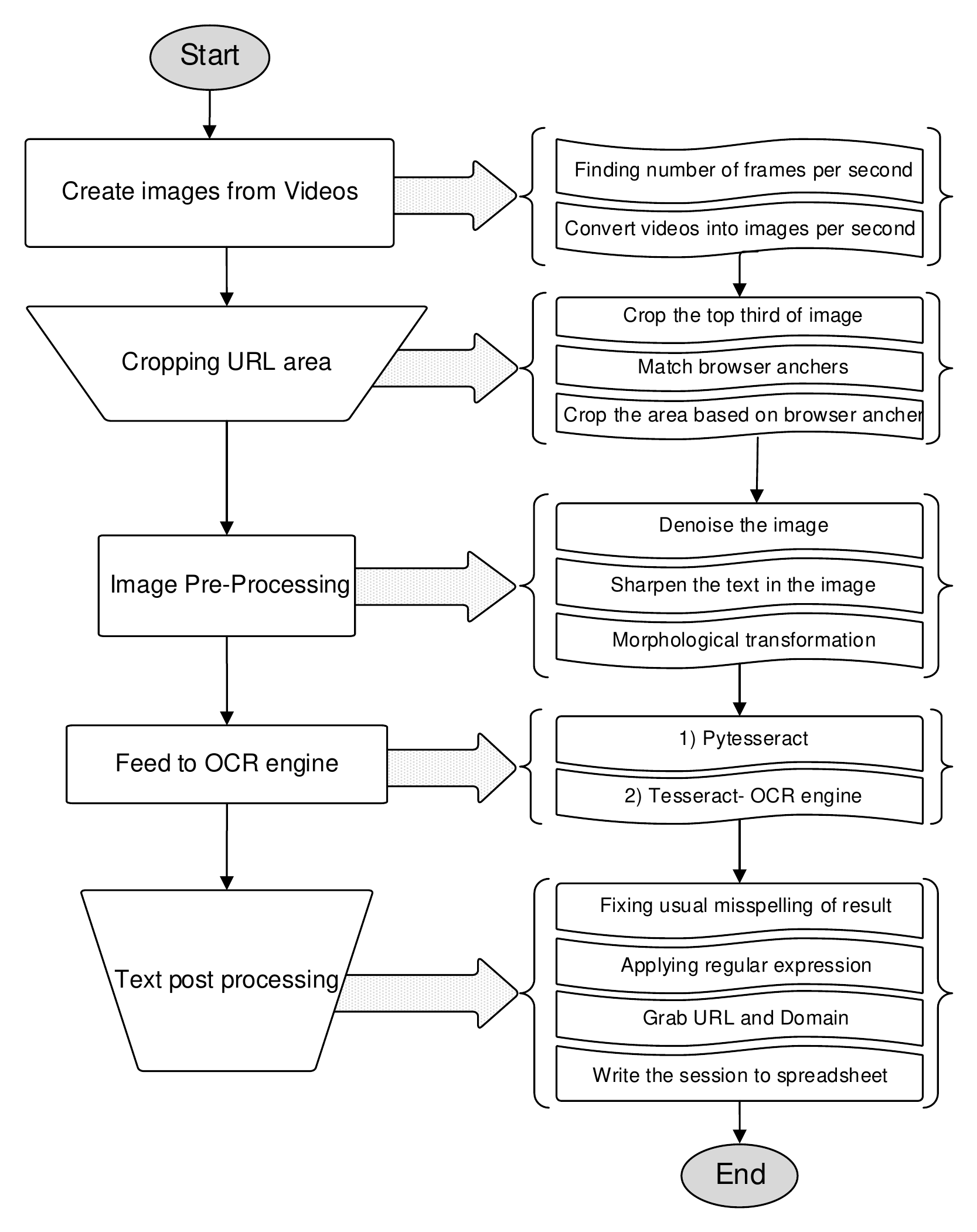}
\caption{General preview of the our implementation steps}
\label{general_view}
\vspace{-20pt}
\end{figure}

\section{Method}\label{Sec.3}
In this section, we discuss the implementation of URL retrieval using OCR, which has been used for this work. First, we describe the general idea and its steps. Later each step will be explained in more detail. The approach we used in this work relies on screen recording of a user's interaction with multiple browsers. Therefore, the input to the pipeline is a video recording or  consecutive screen shots of a computer screen. Free tools such as OBS studio (Windows) and Build-in quick player (Mac) are able to record computer screens as users are interacting with their machines. Also, software such as PC screen capture captures screen shots per second of the online searching sessions of an experimental study. Figure~\ref{general_view} shows the general picture of our implementation structure and steps.\\ The rest of this section describes each of steps shown in Figure~\ref{general_view}.

\subsection{Extracting Images}
In order to retrieve the URLs, first we needed to convert videos into images taken per second. One could easily calculate the correct number of images to generate by considering the frame per second (fps) rate at which the video is recorded. In this work, python Open CV library (CV2) was used to convert the videos to images. The last step was to crop these images so that the template matching method would have a lower chance of picking an area by mistake. We made the assumption that at any given time the URL text field area would be in the top one third of each screen shot. Therefore, we could crop the top $1/3$ of images and discard the rest safely.
\subsection{Template matching}
After retrieving the images per second, we needed to narrow down and specify the URL area for future steps. Our thinking was that doing so would both make it easier for the OCR engine to convert characters into text in less time and produce less noisy data from which to grab URLs and domains later on. Consequently, we needed to define specific anchors around the URL text field which could be detected with template matching algorithms~\cite{opencv:book}. These templates were carefully selected based on the type of browser used for interaction. Python Open CV includes 6 types of template matching algorithms. In this work, we used the TM\_CCOEFF\_NORMED method to perform the template matching using a threshold of 80\% for maximum value of the match. This threshold might need to be altered for other applications depending on the quality of images and fine tuning may  be needed to get the best results. Considering the source image as $I$ and Template image as $T$ and Result matrix as $R$, normalized correlation coefficient matching can be computed by equation~\ref{t_mach1}. 
\begin{equation}\label{t_mach1}
  R(x,y)= \frac{ \sum_{x',y'} (T'(x',y') \cdot I'(x+x',y+y')) }{ Z(x,y) }
\end{equation}


Where 

\begin{equation}
    T'(x',y')=T(x',y') - \frac{1}{(w \cdot h) \cdot \sum _{x'',y''} T(x'',y'')}
\end{equation}

\begin{equation}
    I'(x+x',y+y')=I(x+x',y+y') - \frac{1}{ (w \cdot h) \cdot \sum _{x'',y''} I(x+x'',y+y'')}
\end{equation}
\begin{equation}
    Z(x,y) = \sqrt{\sum_{x',y'}T'(x',y')^2 \cdot \sum_{x',y'} I'(x+x',y+y')^2}
\end{equation}
$w$ and $h$ are the width and height of the template image and Z addresses the normalization part of this algorithm.
Finding a match will provide the height of the anchor which corresponds to the upper corner of browser text field. Using the height of the best match and width of template, we can crop an image such that mostly only the address bar remains. The resulting cropped image undergoes several image processing techniques which will be described in the following section.

\subsection{Image pre-processing}
Depending on the quality of video or screen shots taken during a session, the browser text field may require pre-processing. The process used in this implementation is as follows:
\begin{enumerate}
  \item Convert RGB images to gray scale images.
  Different algorithms exist to convert an RGB image into its corresponding gray scale one. The average method simply averages over all three channels values while the lightness method averages only maximum and minimum vales of all channels.  Open CV library uses the luminosity averaging technique where specific weights would be applied to each channel as given by equation \ref{rgb_scale}
  \begin{equation}\label{rgb_scale}
Gray~scale~value:
  \quad Y = 0.299 \cdot R + 0.587 \cdot G + 0.114 \cdot B   
  \end{equation}
  
  \item Re-scaling image to a bigger size. Although we cannot change the quality of pictures, this step increases the number of pixels and thus makes it possible to improve the result by using other image processing filters. Our experiments showed that re-sizing the image three times of the original yielded the best results.\\
  \item De-noise the resulting images.
  De-noising images is a significant task in image processing and different algorithms has been proposed to effectively perform the task. One can separate these algorithms into three main categories: special domain methods, transform domain methods, and learning based methods~\cite{shao:liu}. We used the Non-local Means Denoising algorithm which comes in the same library to have a consistent implementation purely in python. The algorithm works by considering a noisy image $v=\{v_{i}| i\in \Omega\}$, the result intensity of a pixel $u_{i}$ can be computed by weighted average of neighboring pixels within certain neighborhood $I$ of that pixel.
  \begin{equation}
      u_{i}= \sum\limits_{j\in I}\omega(i,j)v_{j}
  \end{equation}
  where these weights can be computed by equation \ref{nml-weight}.\vspace{0.25cm}
 \begin{equation}\label{nml-weight}
  \omega(i,j) = \frac{1}{\sum_{j}\omega(i,j)} \exp\left( \frac{\| v_{N_{i}}-  v_{N_{j}} \|^{2}_{2,a}}{h^{2}} \right) 
 \end{equation}
 \vspace{0.25cm}
 such that $N_{i}$ refers to the patch size centered at $i$ and $a$ represents the standard deviation of a gussian kernel~\cite{Buades}. Generally, the method averages over all similar pixels and the similarity is measured by comparing patches of the same size around pixels in the search window. To perform denoising in this work, patch size and search windows of 7 and 21 has been selected respectively. These parameters along with filter strength of 10 produced the best result for our experiments.\\
  \item Sharpen the final images.
 To perform sharpening, kernel 2D filter was used. This kernel goes over the original image pixels applying it on windows around that pixel. Figure~\ref{sharpen} shows an example of this process.
\begin{figure}[t]
\centering
\includegraphics[width= \textwidth]{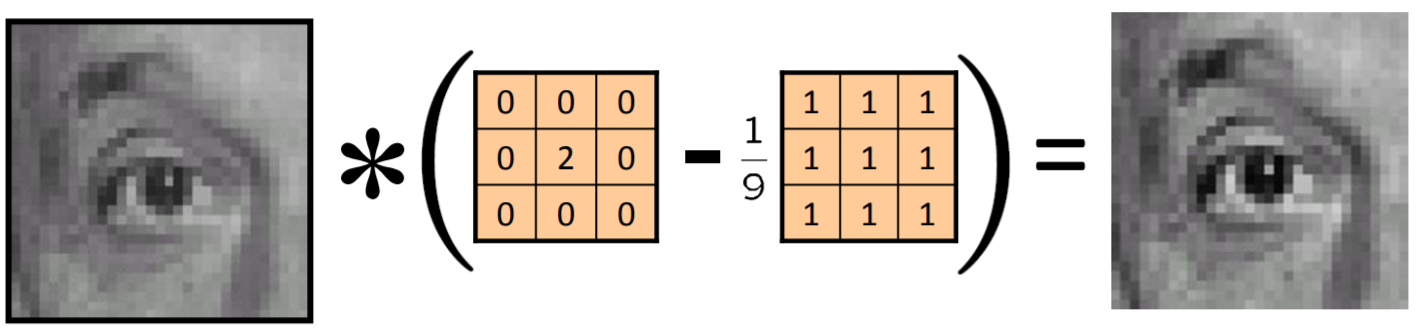}
\caption{Example of applying kernel filter to achieve sharpened image~\cite{cornell}}
\label{sharpen}
\end{figure}
  To perform the sharpening in this work, we applied a 5X5 Gaussian kernel similar to the above picture. Performing the last two steps improves the shape and contrast of the characters with its white surrounding and thus improving the accuracy of OCR engine output.
\end{enumerate}

\subsection{Optical Character Recognition}
The main technology used to find URLs and domains visited by a user is Optical Character Recognition (OCR). Different OCR engines are available but in order to have a consistent open-source solution, we chose to work with Tesseract-OCR as the main engine for character recognition. Tesseact is an open-source OCR engine written in C and C++ in Google~\cite{smith}. Despite being open-source, Tesseact performs well even in comparison with commercial OCR engines such as Transym OCR~\cite{patels}. Moreover, there exists a python wrapper library "pytesseract"  which provides direct access to this engine in python . As a result, the implementation will be completely in python.
Tesseract OCR architecture will be briefly explained here.

\begin{figure}[t]
\centering
\includegraphics[width=\textwidth]{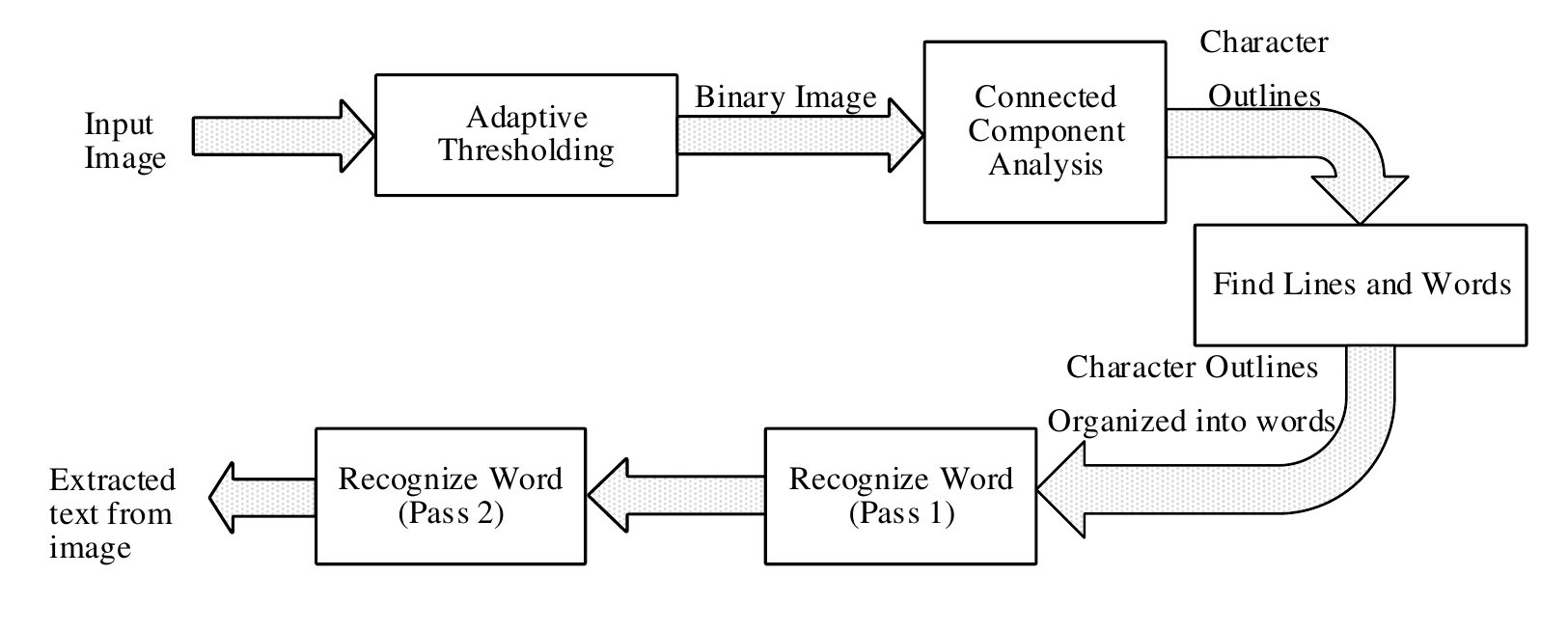}
\caption{Architechture of Tesseract OCR }
\label{tesseract}
\end{figure}

Figure~\ref{tesseract} shows Tesseract architecture and steps that were performed.
As a first step. the image was converted to a binary image by applying an adaptive threshold. Next, using connect component analysis, character outlines were extracted. The outlines then were converted into Blobs which themselves were converted into text lines. Text lines were analyzed for fixed patches and broken down into words using character spacing. Words were further broken into character cells immediately and proportional text was broken using definite space and fuzzy space. Tesseract uses a 2-stage word recognition at the end to improve its result. The satisfactory result of the first pass was given to a classifier as training data to increase the result accuracy.

\subsection{Text Post Processing}
OCR engines try to convert any visual clue in the given image into a character. This leads to unwanted characters appearing in the results. For example, websites that use Hypertext Transfer Protocol Secure (HTTPS) usually include a lock sign in the browser next to their URL text field. Other cases include when the browser is not in full-screen mode and thus, the computer desktop's texts or other browsers' texts might be fed into OCR. OCR tries to convert these shapes into characters that will affect the result. Thus, we needed to use post processing on the result of OCR engine. Regular expressions (Regex) is a powerful tool that took care of these cases. Python regular expression library (re module) provided us access to this tool. Regex can also be used to fix typos and unify the results (e.g., dropping www from the URL if it exists, etc).

\section{Experimental results}\label{Sec.4}
We used the above mentioned approach on the results of a lab experiment where participants were instructed to look at sensitive topics in order to investigate the pathways they used to find their topic of interest, given their religious and political backgrounds. Due to the nature of experiment, participants were given the chance to use a regular web browser or Tor browser to perform their search (all sessions were conducted in a secure research lab). A video (or consecutive screen shots) were recorded from each computer screen during the online search sessions. These videos were stored for later analysis after the session.

\begin{figure}[t]
\centering
\vspace{-0.5cm}
\includegraphics[width= \textwidth]{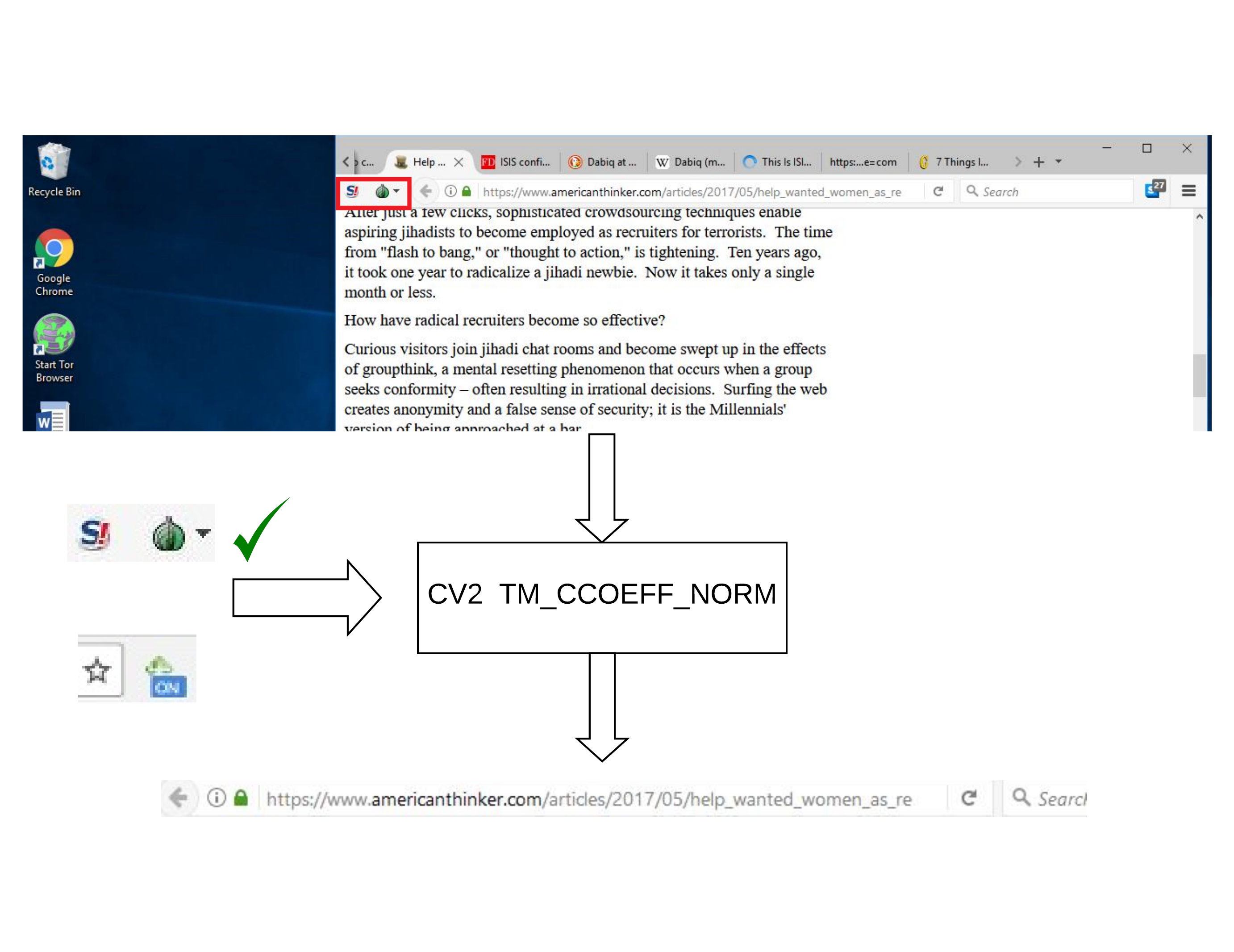}
\vspace{-1.5cm}
\caption{Template matching with anchors specified to Tor and Chrome}
\label{temp_match}
\end{figure}
\begin{figure}[h]
\centering
\includegraphics[width= 8cm]{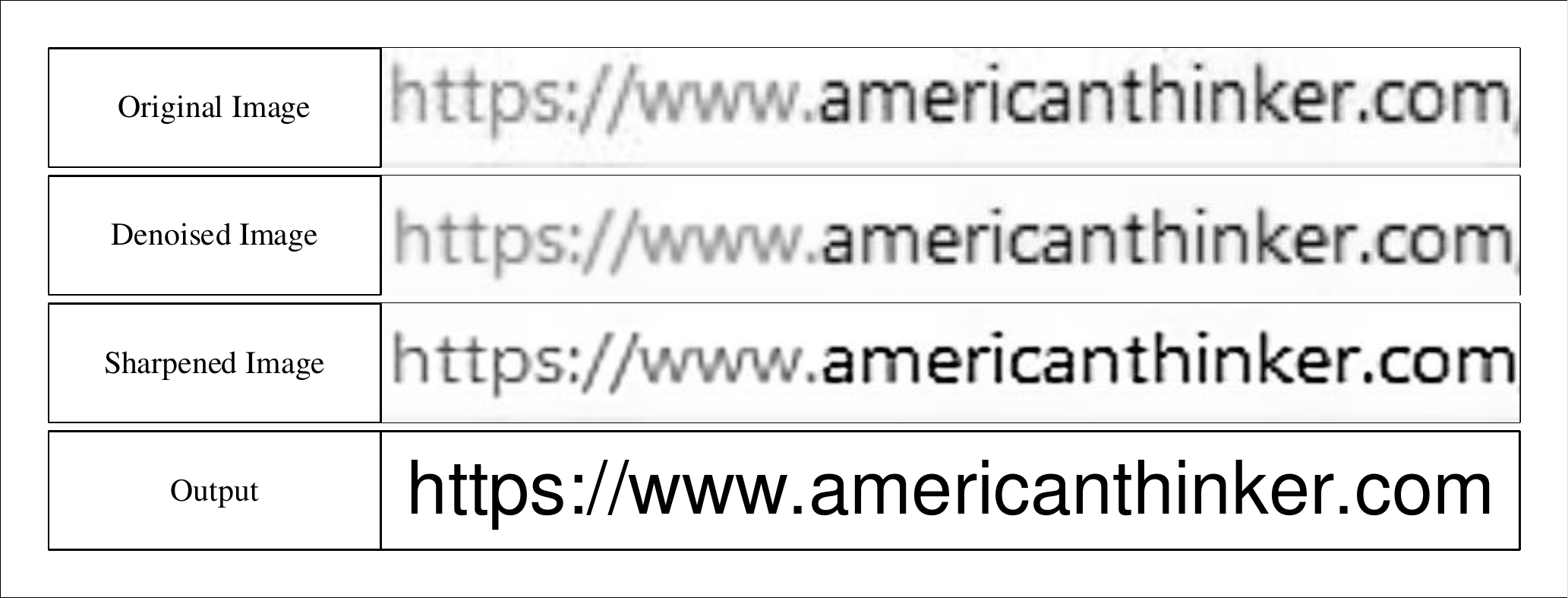}
\caption{Image processing steps illustration used to improve results}
\label{image_output}
\end{figure}

Following the steps explained in section~\ref{Sec.3}, we created a set of images and cropped the top 1/3 of the image. In the experiments we ran, the user could either use  Google Chrome or Tor for browsing the web, we defined the anchors for these two browsers as shown in Figure~\ref{temp_match}.
We used the onion shape in the Tor browser and one extension added to Chrome browser. Depending on the match, the URL text field will be cropped from the image.

\begin{figure}[t]
\centering
\includegraphics[width= 0.9\textwidth]{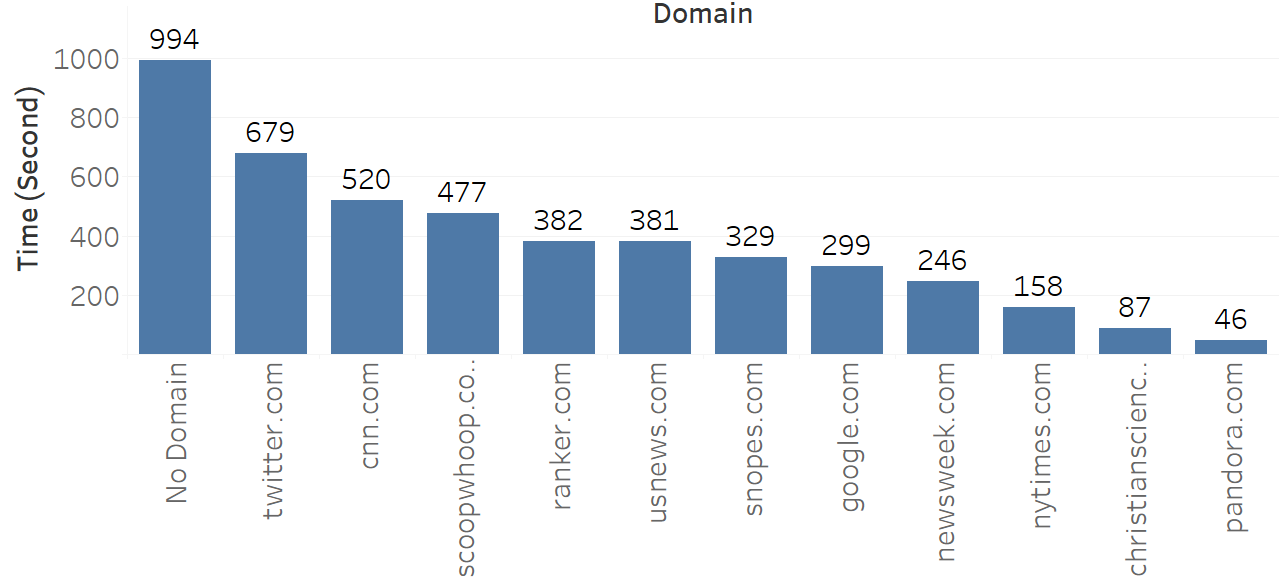}
\caption{Distribution of domains and the time (in second) for a participant}\label{fig:general domain}
\vspace{-0.6 cm}
\end{figure}

The text field was gray-scaled, re-sized, de-noised,and finally sharpened. The final result was fed into Tesseract OCR to extract the text. Figure~\ref{image_output} shows an example of detecting a domain. Finally, text post-processing was performed on the output as needed. 
Collecting the aggregated results of images, we were able to understand which domains participants spent most of their time on, as can be seen in figure~\ref{fig:general domain}.\\

\begin{figure}[b]
\centering
\includegraphics[width= \textwidth]{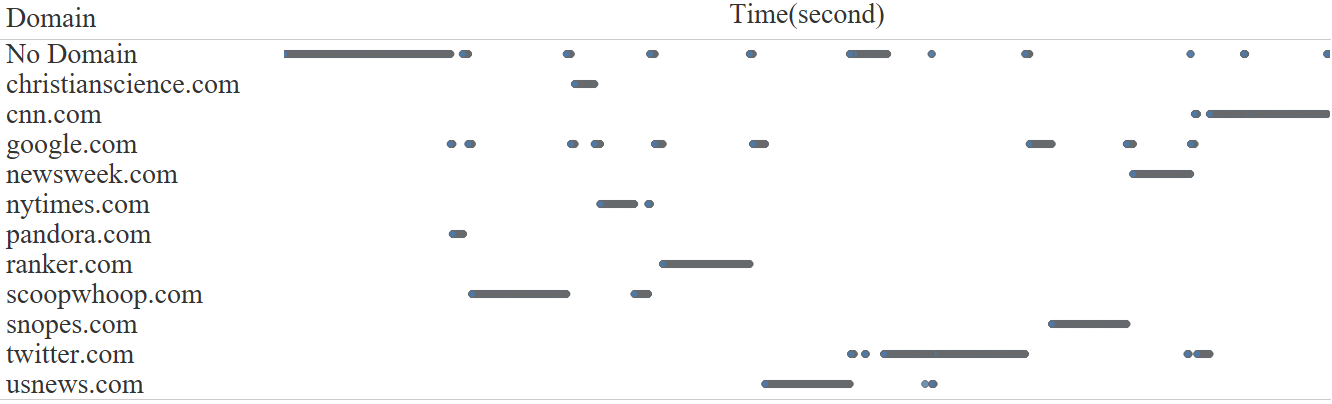}
        \caption{The path that the participant took in the same experiment. Horizontal axes represent the time passed of the session}
        \label{fig:path}
\end{figure}

Using this process, we can investigate the paths each participant took during their interaction with the internet regardless of the type of browser they chose to use. Figure~\ref{fig:path} shows a visualization of the same user path along with the duration of the experiment that had passed. This visualization helps to understand where a particular participant starts and ends, given their interests and backgrounds.

\section{Conclusion and Future works}\label{Sec.5}
This paper presents a different approach for tracking user behaviours as they interact with the World Wide Web (WWW). We used image processing methods along with an OCR engine (tesseract-OCR) to extract URLs and domains visited by a participant in a session. This was particularly difficult due to the nature of these experiments where each participant was allowed to use multiple browsers (Chrome or Tor) as the web browsing tool and could switch between them at will. The computer screen and audio recording during the interaction session was kept for separate analysis later on. To extract visited domains, we converted videos into consecutive images (if needed) and cropped only the URL area of each browser using template matching. The resulting area went into image pre-processing and then fed into the OCR engine and finally using regular expressions, we extracted the URL and domain within each image.

With very few exceptions, by using this method we were able to collect correct URLs and track the path a participant took during the experiment. This was mainly due to the quality of videos and images causing some noise into pictures that was extremely difficult to remove. However, since most of these pictures were taken consecutively, one could find the correct domain by checking other very similar results extracted from other frames. As for future research, one could try to automate such cases as well. This could be done by sending requests to these domains and replace the similar non-perfect results with domains that correctly return the requests. Moreover, this approach could be used to extract the content of a web page as well as the advertisements around the page depending on their importance in experiments. Lastly, this methodology could aid in the effort for machines to learn how to identify users' patterns of behaviour online and respond to potential online threats.

\bibliographystyle{bibtex/spmpsci} 
\bibliography{ref}

%
%

\end{document}